\documentclass[11pt,a4paper]{article}
\usepackage[]{authblk}
\usepackage[hyperref]{acl2021}
\usepackage{times}
\usepackage{latexsym}
\usepackage{graphicx}
\usepackage{amsmath}
\usepackage{multirow}
\usepackage{todonotes}
\usepackage{xspace}
\usepackage{caption}
\usepackage{subcaption}
\usepackage{booktabs}

\usepackage{microtype}

\newcommand*{\argmax}{\ensuremath{\operatornamewithlimits{argmax}}\xspace}

\newcommand\BLEU{\textsc{Bleu}}

\newcommand\pst{p_\text{s2t}}
\newcommand\psp{p_\text{s2p}}
\newcommand\ppt{p_\text{p2t}}

\newcommand\Encoder{\text{Encoder}}
\newcommand\Decoder{\text{Decoder}}

\aclfinalcopy 

 \title{
Integrated Training for Sequence-to-Sequence Models Using Non-Autoregressive Transformer 
}
\author[1,2]{Evgeniia Tokarchuk}
\author[2]{Jan Rosendahl}
\author[2]{Weiyue Wang}
\author[1]{Pavel Petrushkov}
\author[1]{Tomer Lancewicki}
\author[1]{\authorcr Shahram Khadivi}
\author[2]{Hermann Ney}
\affil[1]{eBay Inc., Aachen, Germany}
\affil[2]{RWTH Aachen University, Aachen, Germany}
\affil[ ] {\texttt{e.tokarchuk@uva.nl}}
\affil[ ]{\texttt{\{ppetrushkov,tlancewicki,skhadivi\}@ebay.com}}
\affil[ ]{\texttt{\{rosendahl,wwang,ney\}@cs.rwth-aachen.de}}

\date{}

\begin{document}

\maketitle
\begin{abstract}
Complex natural language applications such as speech translation or pivot translation traditionally rely on cascaded models. 
However, cascaded models are known to be prone to error propagation and model discrepancy problems.
Furthermore, there is no possibility of using end-to-end training data in conventional cascaded systems, meaning that the training data most suited for the task cannot be used.
Previous studies suggested several approaches for integrated end-to-end training to overcome those problems, however they mostly rely on (synthetic or natural) three-way data.
We propose a cascaded model based on the non-autoregressive Transformer that enables end-to-end training without the need for an explicit intermediate representation.
This new architecture (i) avoids unnecessary early decisions that can cause errors which are then propagated throughout the cascaded models and (ii) utilizes the end-to-end training data directly.
We conduct an evaluation on two pivot-based machine translation tasks, namely French$\to$German and German$\to$Czech.
Our experimental results show that the proposed architecture yields an improvement of more than 2 \BLEU{} for French$\to$German over the cascaded baseline.
\end{abstract}

\section{Introduction}
Many complex natural language applications such as speech translation~\cite{sperber-paulik-2020-speech} or pivot translation~\cite{utiyama-isahara-2007-comparison, de2006catalan} traditionally rely on cascaded models.
The technique of model cascading is commonly used to solve problems that can be divided into a sequence of sub-problems where the solution to the first problem is used as an input to the second and so on.
Typically cascaded systems include several consecutive and independently trained models, each of which aims to solve a particular sub-task.
For example in a cascaded speech translation system an automatic speech recognition model receives the audio signal as an input and generates a transcription as an output of the first sub-task.
This output could be passed to a system that adds punctuation and capitalization to the sequence, before, as a final step, a machine translation system is applied.

Cascaded models are appealing if there is more training data for each of the sub-tasks than for the full task.
Examples for such scenarios include automatic speech translation (AST), image captioning in non-English languages, and non-English machine translation. However, cascaded models are prone to error propagation, meaning that decision errors in the first model are forwarded to and possibly amplified by the second model.
Usually, there is also a loss of information when passing information between models since the interface between models traditionally requires each model to output a discrete decision.
This means that the deeper knowledge that the model may encode in its representation of the output is reduced to a \lq surface form\rq{} of a particular prediction, which is passed on to the following model.
Lastly, in conventional cascaded system there is no possibility to make use of end-to-end training data, meaning that the training data most suited for the task cannot be used.

To tackle these problems, several approaches for integrated end-to-end training of cascaded models have been proposed and applied to different NLP tasks~\cite{bahar2021:slt:st,sperber-etal-2019-attention,Sung-2019-end-to-end-speech}.
Integrated end-to-end training is usually achieved by merging the consecutive models and fine-tuning the resulting system on the end-to-end training data.
Although the idea of this approach is simple, it remains an open challenge how to choose the interface between the models in such a way that they can be trained, e.g. by gradient propagation.
Furthermore, most of these approaches rely on synthetic or natural multi-way training data, i.e. data that does not only provide an $(\text{input}, \text{output})$ pair but also the correct label for all sub-tasks involved.
For a detailed discussion of the literature, we refer to Section~\ref{sec:related-work}.
In this work we focus on the task of pivot-based machine translation, i.e. the translation from a source (src) language via a pivot (piv) language to the desired target (trg) language, as an example for a two-stage task that is traditionally solved by model cascading.


We propose a cascaded model based on the non-autoregressive Transformer (NAT) that enables end-to-end training without the need for an explicit intermediate representation, that is inevitable in autoregressive models.
This new architecture (i) avoids unnecessary early decisions that can cause errors which are then propagated throughout the cascaded models (ii) utilizes the src$\to$trg, src$\to$piv and piv$\to$trg training data and (iii) communicates the full information from the src$\to$piv model downstream by providing a natural interface between the src$\to$piv and piv$\to$trg models.





\section{Related Work}
\label{sec:related-work}
Several approaches were proposed in recent years to address the weaknesses of the traditional cascaded models. Early works investigated the applications of the N-best list decoding both in speech translation and pivot-based translation~\cite{Woszczyna-1993-nbest,lavie-etal-1996-nbest,och-ney-2004-alignment,utiyama-isahara-2007-comparison}. The N-best list decoding allows to pass multiple intermediate hypotheses and avoid unnecessary early decisions. An efficient alternative to the $n$-best list is lattices, which replaced the $n$-best list for the speech translation models~\cite{Zhang2005-lattice,Schultz2004-lattice, matusov08:interspeech}. However, the usage of the discrete decisions does not allow to train cascaded model jointly on src$\to$trg data. 

Most recent works are focusing instead on the joint or integrated training for sequence-to-sequence cascaded models. Thus, ~\cite{cheng-2017-jointpivot} suggested a joint training approach for the pivot-based neural machine translation. In their work, two attention-based RNN models~\cite{BC2015} are trained jointly with different connection terms in the objective function and the src$\to$trg as a bridging corpus.
Another approach is to apply the transfer-learning technique for pivot-based NMT~\cite{kim-2019-pivot-transfer}, meaning that the direct src$\to$trg model is initialized with the respective weights from the pre-trained models, and fine-tuned on src$\to$trg corpus through the trainable adapter.
Pivot-based NMT is typically used in a low-resource src$\to$trg setup, and multilingual NMT systems proved to be successful in this scenario~\cite{johnson-etal-2017-googles-mlt, aharoni-etal-2019-massively-mlt,zhang-etal-2020-improving}.
To tackle a low-resource NMT problem,~\cite{kim-2019-pivot-transfer} also explore different ways to extend the back-translation idea~\cite{sennrich-etal-2016-bt} for src$\to$piv$\to$trg scenarios. However, since this work aims to provide the general framework for the integrated training of cascaded sequence-to-sequence models, we do not aim for comprehensive comparisons with multilingual NMT systems and various data augmentation strategies. We refer to~\cite{kim-2019-pivot-transfer} for in-depth comparison studies.

In speech translation, the tight model integration for the cascaded models also attracted attention from the community.  ~\cite{anastasopoulos-chiang-2018-tied,wang-etal-2019-bridging,sperber-etal-2019-attention} discussed either use of attention or hidden state vectors as a connection interface for the tight model integration in cascaded systems. Recently, ~\cite{bahar2021:slt:st} proposed to use posterior distribution as an input to the encoder of the second model.

\section{Background}
\subsection{Sequence-to-Sequence modeling}
The modeling of the sequence-to-sequence problems, namely converting the source sequence $f_1^J$ in one domain to the target sequence $e_1^I$ in another domain, is nowadays usually done using encoder-decoder deep neural networks~\cite{SV2014,BC2015,Vaswani-2017-attention}. The purpose of the encoder is to map the input sequence $f_1^J$ to a continuous, hidden vector representation $h$, from which the decoder decodes the target sequence. 

In applications such as machine translation, the Transformer~\cite{Vaswani-2017-attention}, an attention-based sequence-to-sequence model, is  considered state of the art~\cite{barrault-EtAl:2020:WMT1}.

Commonly the probability distribution over the target sequences in sequence-to-sequence models is expressed by a left-to-right factorization:
\begin{align}
\label{eq:ar-dep}
            p(e_1^I|f_1^J) = \prod_{i=1}^I p(e_i|e_1^{i-1}, f_1^{J}).
\end{align}
These models are also called \textit{autoregressive}, meaning that each consecutive token in the target sequence depends on the left context of the same sequence.

\subsection{Non-Autoregressive NMT}
In contrast to the autoregressive modelling approach, the non-autoregressive Transformer~\cite{GB18} assumes that all tokens in the target sequence are generated independently of each other.
This means in particular that there is no need for a search procedure at inference time since target tokens can be generated and optimized in parallel.
However, current approaches also need an explicit length model as additional input to the decoder.
\citet{GB18} utilize the standard Transformer architecture and provide several modifications in order to obtain a non-autoregessive machine translation system. 

Recent works proposed to relax the independence constraints during training and use \textit{iterative decoding} for the NAT, meaning that instead of only one decoding pass, the model relies on the multiple passes, and conditional dependence might be used on the consecutive passes to achieve better performance~\cite{ghazvininejad-etal-2019-mask,gu-2019-levenshtein, lee-etal-2018-deterministic, stern-2019-insertion-transformer}. Such decoding procedure allows shrinking the gap between the performance of the autoregressive and non-autoregressive models. 

\subsection{Pivot-based Machine Translation}
\label{sec:pivot-tr}
A cascading system $\pst$ for pivot-based machine translation consists of a src$\to$piv model $p_\text{s2p} $ and a piv$\to$trg model $p_\text{p2t} $, which 
typically have a disjoint parameter set.
While both models are trained independently, they work in cooperation when producing the translation, i.e., the most likely target sequence $\hat{e}_1^{\hat{I}}$ for the given source sequence  $f_{1}^{J}$. 
The pivot sequence $z_1^K$ can be viewed as a latent variable, and the target sequence probability can be expressed by summing over all pivot sequences:
\begin{align*} 
          \pst(e_1^I|f_1^J) 
          &= \sum_{z_1^K}{\ppt(e_1^I|z_1^K,f_1^J) \psp(z_1^K|f_1^J)} \\
          & = \sum_{z_1^K}{\ppt(e_1^I|z_1^K) \psp(z_1^K|f_1^J)}.
\end{align*}
Since the sum over all possible pivot hypothesis $z_1^K$ is intractable in practice, instead \textit{two-pass decoding} is used as an approximation to obtain the target hypotheses:
\begin{align}
\label{eq:two-pass-decoding}
            \hat{z}_1^{\hat{K}} &=  \argmax_{K, z_1^K}{\prod_{k=1}^K \psp (z_k|z_1^{k-1}, f_1^J)} \\
            \hat{e}_1^{\hat{I}} &= \argmax_{I, e_1^I}{\prod_{i=1}^I \ppt (e_i|e_1^{i-1}, \hat{z}_1^{\hat{K}})}.
\end{align}
We investigate the stability and potential for improvement of this interface in the Section~\ref{sec:eror-propagation}.

\section{Model Integration}
\label{sec:model-integration}

\begin{figure*}[ht]
    \centering
    \begin{subfigure}[b]{0.45\textwidth}
        \centering
         \includegraphics[width=0.7\textwidth]{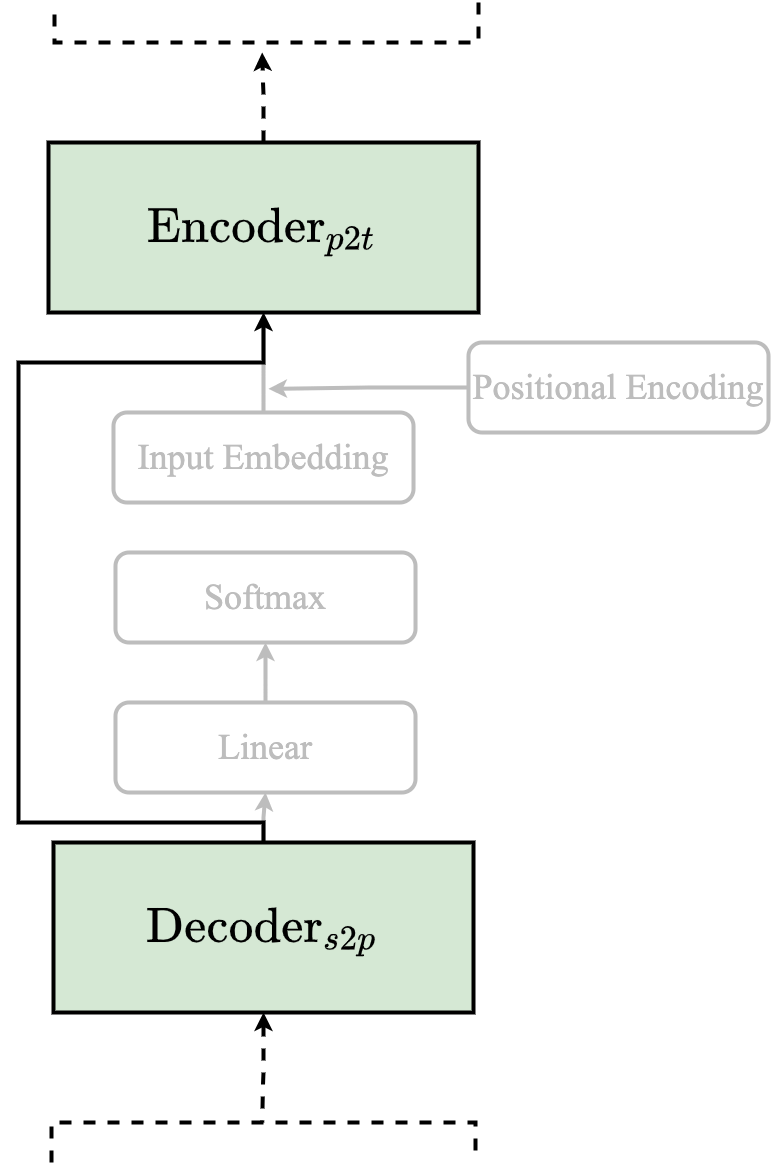}
         \caption{Decoder States Interface.}
         \label{fig:hs-int}
    \end{subfigure}
    \hfill
    \begin{subfigure}[b]{0.45\textwidth}
        \centering
         \includegraphics[width=0.7\textwidth]{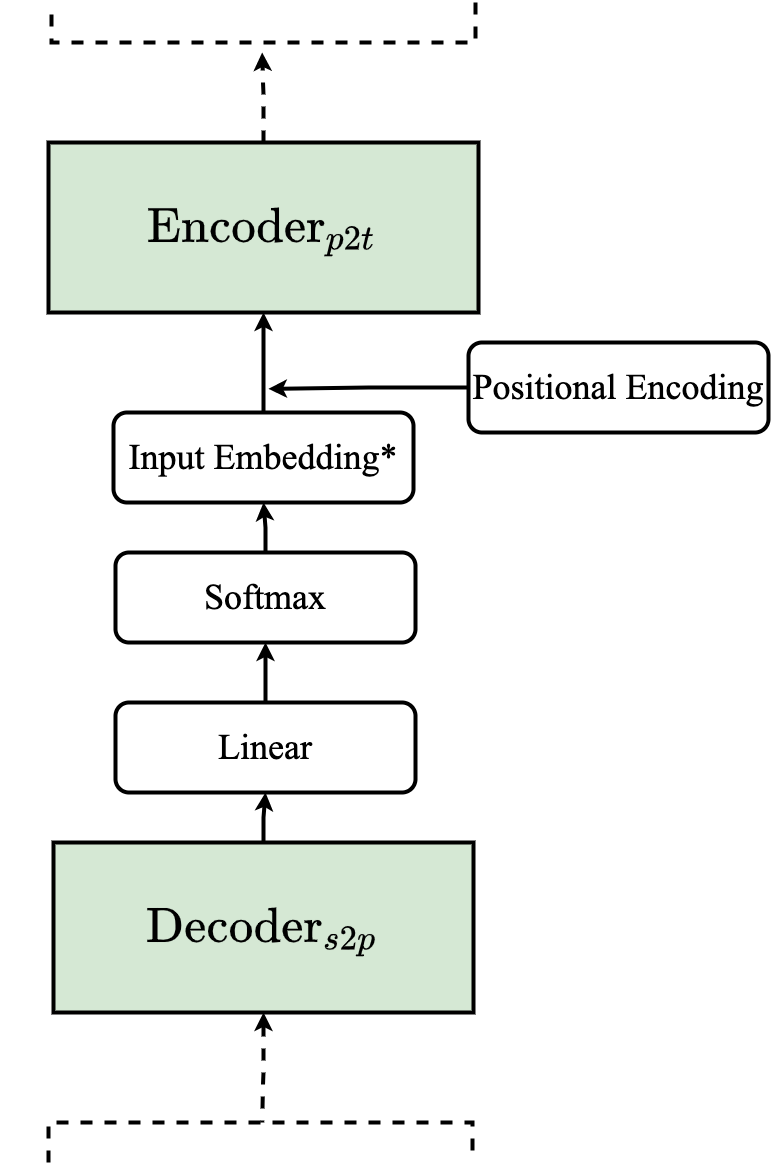}
    \caption{Decoder Posteriors Interface.}
    \label{fig:sm-int}
    \end{subfigure}
    \caption{Two proposed connection interfaces between src$\to$piv and piv$\to$trg models for integrated training. The blocks in gray represents are omitted layers of the original cascaded Transformer architecture.
    For simplicity we do not show the $\Encoder_{s2p}$ and $\Decoder_{p2t}$. \\
    *Note that the input embedding is now a full fledged matrix multiplication, not a multiplication with a one-hot vector which is equivalent to a column selection.
    }
    \label{fig:intefaces}
\end{figure*}
Starting from the conventional cascaded model, as described in Section~\ref{sec:pivot-tr}, we propose to connect the two consecutive encoder-decoder models through an end-to-end trainable interface.
The src$\to$piv model consists of both $\Encoder_{s2p}$ and $\Decoder_{s2p}$, similarly the piv$\to$trg model consists of $\Encoder_{p2t}$ and $\Decoder_{p2t}$. We introduce an interface which connects $\Decoder_{s2p}$ to the $\Encoder_{p2t}$. The main requirement for this connection interface is to be differentiable to make the gradient propagation possible. In order to fulfill this requirement, we follow the previous work (see more in Section~\ref{sec:related-work}) and choose to focus on two possible interfaces:

\begin{itemize}
    \item \textbf{Decoder States Interface}: Pass the final sequence of hidden states vectors of the last src$\to$piv $\Decoder_{s2p}$ layer as an input to the $\Encoder_{p2t}$.
    The input embedding layer and positional encoding layer are omitted in the $\Encoder_{p2t}$, and the hidden states vector is then used directly as an input to the next self-attention block (see Figure~\ref{fig:hs-int}).
    \item \textbf{Decoder Posteriors Interface}: Pass the probability distribution  $p_{s2p}(z_1^K|i,f_1^J)$ of the $\Decoder_{s2p}$.
    The embedding matrix $E$ from $\Encoder_{p2t}$ is used to calculate a \lq soft embedding\rq{} 
    \begin{equation*}
        \sum_{v\in V} E_v p_{s2p}(z_k=v|f_1^J).
    \end{equation*}
    Hence, the $\Decoder_{s2p}$ and $\Encoder_{p2t}$ are connected through the softmax layer, as shown shown in Figure~\ref{fig:sm-int}.
\end{itemize}
\begin{figure*}[ht]
    \subfloat[AR-based integrated model.\label{fig:ar-int}]{%
        \centering
         \includegraphics[width=1\linewidth]{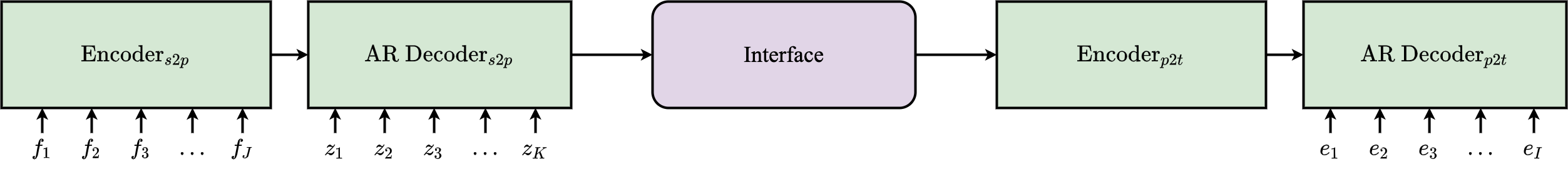}%
    }\par\vspace{0.3cm}
    \subfloat[NAT-based integrated model.\label{fig:nat-int}]{%
        \centering
         \includegraphics[width=1\linewidth]{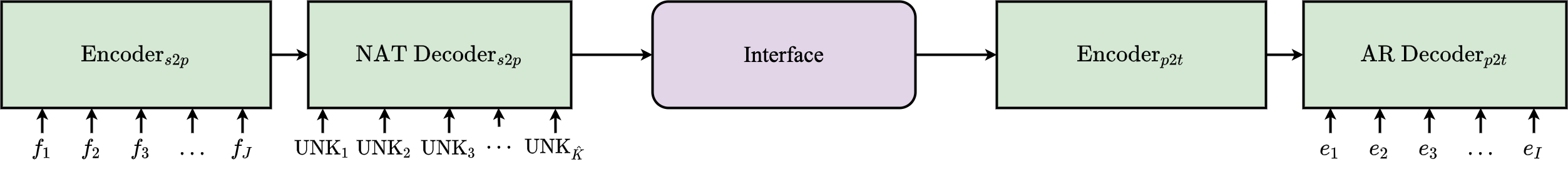}%
    }\par\vspace{0.3cm}
    \subfloat[Three-components NAT-based integrated model. \label{fig:nat-int-small}]{%
        \centering
         \includegraphics[width=1\linewidth]{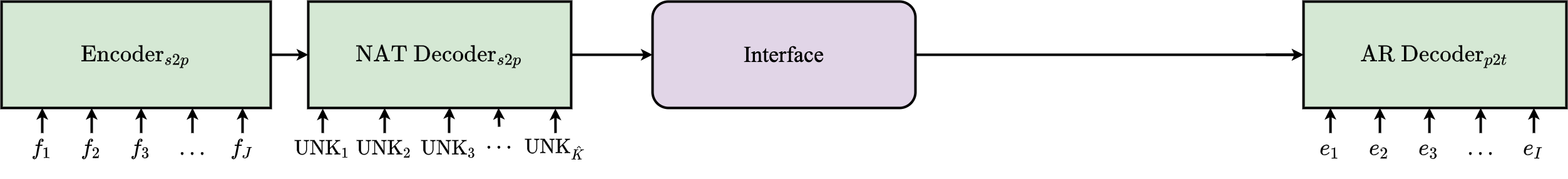}%
    }
 \caption{Different variants of the encoder-decoder model integration through the connection interface.} 
 \label{fig:arch}
\end{figure*}
Note that the decoder posteriors interface requires the src$\to$piv and piv$\to$trg model to share a common vocabulary~$V$.

Two autoregressive encoder-decoder models can be connected through these interfaces as shown in Figure~\ref{fig:ar-int}. However, at training time the $\Decoder_{s2p}$ requires a pivot sequence as an input.
If there is no access to the three-way src$\to$piv$\to$trg data, the pivot sequence has to be obtained by doing a search in training, which is computationally very prohibitive in a real world task, or via forward or backward translation beforehand (synthetic data).
The disadvantage of using synthetic data is that the pivot sequences remain static throughout the training, this means that the cascaded src$\to$piv$\to$trg model is trained on pivot sequences which become less relevant the more training updates the src$\to$piv models receives.
To avoid a sub-optimal, discrete intermediate representation while still benefit from the model integration, we propose to replace src$\to$piv autoregressive Transformer with a non-autoregressive one as shown in Figure~\ref{fig:nat-int}.
The usage of NAT allows to replace the pivot sequence with a sequence of unknowns during the training on src$\to$trg data.
Since the decoder states interface do not use the embeddings of the $\Encoder_{p2t}$, similar to other works, the $\Encoder_{p2t}$ can be safely omitted in the integrated model (Figure ~\ref{fig:nat-int-small}).

Training such a cascaded model can be done with the following steps:
\begin{itemize}
    \item \textit{Pre-training}:
    \begin{itemize}
        \item Train src$\to$piv model on src$\to$piv corpora
        \item Train piv$\to$trg model on piv$\to$trg corpora
    \end{itemize}
    \item \textit{Concatenation}: Concatenate the models in the cascade through the interface and initialize respective components with the pre-trained weights.
    \item \textit{Fine-tuning}: fine-tune the resulting integrated model on the src$\to$trg data.
\end{itemize}
This yields a src$\to$trg architecture in which all parameters are pre-trained and which makes use of all parameters from the pre-trained models, with the exception of one linear layer and an embedding matrix in the decoder states interface.
Please note that although we are focusing on pivot-based NMT as our target task, we argue that the proposed integration method can be easily adapted to any Transformer-based cascaded model.

\section{Experimental Results}
To test and verify the proposed cascaded model, we conduct experiments on French$\rightarrow$German and German$\rightarrow$Czech data from the WMT 2019 news translation task\footnote{\url{http://www.statmt.org/wmt19/}}. 

\subsection{Data}
Training data for French$\rightarrow$German includes Europarl corpus version 7~\cite{koehn2005europarl}, CommonCrawl\footnote{\url{https://commoncrawl.org/}} corpus and the newstest2008-2010. The total number of parallel sentences is 2.3M. 

The original German$\rightarrow$Czech task was constrained to unsupervised translation, but we utilized the available parallel data to relax these constraints. The corpus consists of NewsCommentary version 14~\cite{TIEDEMANN12.463} and we extended it by including newssyscomb2009\footnote{\url{http://www.statmt.org/wmt09/system-combination-task.html}} and the concatenation of previous years test sets newstest2008-2010 from the news translation task. The total amount of parallel sentences is 230K. 

For both tasks we use newstest2011 as the development set and newstest2012 as the test sets.
The data statistics, including pre-training data, are collected in Table~\ref{tab:data-stats}.

\begin{table}[ht]
    \centering
    \resizebox{\linewidth}{!}{%
    \begin{tabular}{llcc}
    \hline
    \multicolumn{2}{l}{}                  & Sentences   & Words (target) \\ \hline
    direct data & French$\rightarrow$German & 2.3M & 53M \\ \hline 
    \multirow[t]{2}{*}{pre-train} & French$\rightarrow$English & 35M & 905M \\ 
    & English$\rightarrow$German & 9.7M & 221M \\ \hline\hline
    direct data & German$\rightarrow$Czech & 230K & 4.5M \\ \hline
    \multirow[t]{2}{*}{pre-train} & German$\rightarrow$English & 9.1M & 180M \\ 
    & English$\rightarrow$Czech & 49M & 486M \\ \hline
    \end{tabular}%
    }
    \caption{Training data overview.}
    \label{tab:data-stats}
    \end{table}
    
\subsection{Preprocessing}
 For each parallel corpus, we apply a standard preprocessing procedure:
 First, we tokenize each corpus using the Moses\footnote{\url{http://www.statmt.org/moses/}} tokenizer. Then a true-casing model is trained on all training data and applied to both training and test data. In the final step, we train \textit{byte-pair encoding} (BPE)~\cite{sennrich-etal-2016-neural}  with 32000 merge operations. In order to enable model integration, we train BPE jointly on all available data for the respective language.

\subsection{Model and Training}
We implement the models described in Section~\ref{sec:model-integration} using the \textit{fairseq}~\cite{ott2019fairseq} sequence-to-sequence extendable framework. As non-autoregressive src$\to$piv model, we choose the Conditional Masked Language Model (CMLM)~\cite{ghazvininejad-etal-2019-mask} with 6 layers for both encoder and decoder, and a standard 6 layer \lq base\rq{} Transformer for the piv$\to$trg system~\cite{Vaswani-2017-attention}.
For each interface, the length of the pivot sequence is set to the length of the source sequence by default. More on the length modeling is discussed in the Section~\ref{sec:length-modeling}. For the decoder states interface, the last decoder is used for all the experiments. 

For model fine-tuning, the Adam optimizer~\cite{kingma2015-adam} with $\beta=\{(0.9,0.98)\}$ and the learning rate $0.5\times10^{-5}$ is used for all the models. The learning rate is reduced during training based on the inverse square root of the update. Additionally, 10,000 and 4,000 warm-up updates have been used for French$\rightarrow$German and German$\rightarrow$Czech accordingly.   
The dropout is set to 0.1 for French$\rightarrow$German and 0.3 for German$\rightarrow$Czech. We set the effective batch size to 65,536 following the fairseq recommendations for the non-autoregressive models. Although CMLM provides the Mask-Predict decoding algorithm~\cite{ghazvininejad-etal-2019-mask}, in our work we only use one iteration and obtain probability distribution and hidden states from the fully masked sequence, which means that each token is only conditioned on the source tokens.
Results are reported using the  \textit{sacreBLEU}\footnote{\url{https://github.com/mjpost/sacrebleu}} implementation of \BLEU{} \cite{papineni-etal-2002-bleu}.

We compare our models against three baselines:
\begin{itemize}
    \item \textit{direct baseline}: The direct baseline is the Transformer base model, which is trained only on src$\to$trg (direct) parallel data.
     \item \textit{AR pivot baseline}: 
    A baseline system composed of cascading a src$\to$piv and a piv$\to$trg autoregressive (AR) models. These two models are autoregressive Transformer `base' models with six layers of encoder and decoder, respectively.
    The individual models are trained on either src$\to$piv or piv$\to$trg data. There is no fine-tuning on the src$\to$trg data, and results are reported based on the inference only.

    \item \textit{NA pivot baseline}: Similarly to the AR baseline, we provide the results for the non-autoregressive (NA) pivot baseline. The main difference is that the non-autoregressive CMLM model is selected as the src$\to$piv model. We follow standard training procedure for the CMLM as described in ~\cite{ghazvininejad-etal-2019-mask}, and as for hyperparameters, we rely on the fairseq guidelines\footnote{\url{https://github.com/pytorch/fairseq/blob/master/examples/nonautoregressive_translation/scripts.md}}.
    While pre-training, a random mask is applied to the decoder input, meaning that the number of observed and masked tokens varies for each batch. During decoding, we employ five decoding iterations to achieve better performance on the src$\to$piv model. The Transformer base piv$\to$trg model is trained in the same way as for the AR pivot baseline.
\end{itemize}


\begin{table*}[ht]
\centering
\begin{tabular}{cllccccc}
\toprule
 & & & \multicolumn{2}{c}{French$\rightarrow$German} && \multicolumn{2}{c}{German$\rightarrow$Czech}\\ \cline{4-5} \cline{7-8}
 & & &\multicolumn{2}{c}{$\BLEU^{[\%]}$} && \multicolumn{2}{c}{$\BLEU^{[\%]}$} \\
& & & dev & test && dev & test\\  \midrule
 \parbox[t]{2mm}{\multirow{2}{*}{\rotatebox[origin=c]{90}{AR}}}& \multicolumn{2}{l}{direct baseline} & 20.0 & 20.4 && 13.5 &  14.0 \\
 & \multicolumn{2}{l}{pivot baseline} &  19.5 & 20.7  && 18.8	& 18.1 \\ \midrule
  \parbox[t]{2mm}{\multirow{4}{*}{\rotatebox[origin=c]{90}{NA Int.}}}
  & \multicolumn{2}{l}{Pivot hypothesis (NA pivot baseline)} & 17.1  & 18.1  && 17.3  & 16.6  \\
 & Decoder States & w/o $\Encoder_{p2t}$  & 20.9 & 21.8 && 15.5 & 15.5\\
 & &  w $\Encoder_{p2t}$  & 21.5 &  22.8 && 16.5 & 16.7 \\
 & \multicolumn{2}{l}{Decoder Posteriors} & 21.6 &22.7 && 16.8 & 17.0\\[0.3em] \midrule  
 \parbox[t]{2mm}{\multirow{2}{*}{\rotatebox[origin=c]{90}{AR Int.}}} 
 & \multicolumn{2}{l}{Decoder States$^{\dagger}$} & 20.6 & 21.2 && 16.6 & 16.8 \\
 & \multicolumn{2}{l}{Decoder Posteriors$^{\dagger}$} & 20.5 & 21.1  &&  17.9 & 17.1 \\[0.4em] 
 \bottomrule
\end{tabular}
\caption{Results for integrated training with different non-autoregressive (NA) interfaces on src$\to$trg data in comparison to autoregressive (AR) baseline model. All pivot/cascaded models are pre-trained on the respective data. We use \texttt{newstest\{2011,2012\}} as dev and test respectively. Results marked with $^{\dagger}$ are taken from ~\cite{hilmes-pivotnmt}.}
\label{tab:results}
\end{table*}
Additionally, we compare our NA integrated model with the AR integrated model (\ref{fig:ar-int}) based on the synthetic data generation~\cite{hilmes-pivotnmt}. Synthetic data is generated by the forward pass of the src$\to$piv model offline before fine-tuning on the src$\to$trg data, meaning that the pivot hypotheses stay the same during fine-tuning.


We report the best results for the proposed cascaded model with the different interfaces in Table~\ref{tab:results}. The best checkpoint is selected based on \BLEU{} score of the development set. The results show up to 2.1\% \BLEU{} improvements for the decoder states and decoder posteriors interfaces on French$\rightarrow$German compare to the pivot baseline. On the other hand, there is a 2.0\% \BLEU{} degradation of the performance while using decoder posteriors interface on German$\rightarrow$Czech compare to the pivot baseline and up to 2.3\% \BLEU{} degradation using decoder states interface. We suppose that such degradation can be based on the training data size since the German$\rightarrow$Czech is ten times smaller than French$\rightarrow$German. To check on our assumption, we perform additional analysis with the different training data partitions in Section~\ref{sec:data-size}. Moreover, according to the decoder states interface results, the usage of the additional encoder showed its usefulness compared to the three-components architecture.

\section{Analysis}
\subsection{Error Propagation}
\label{sec:eror-propagation}
\begin{figure}
    \centering
    \includegraphics[width=\linewidth]{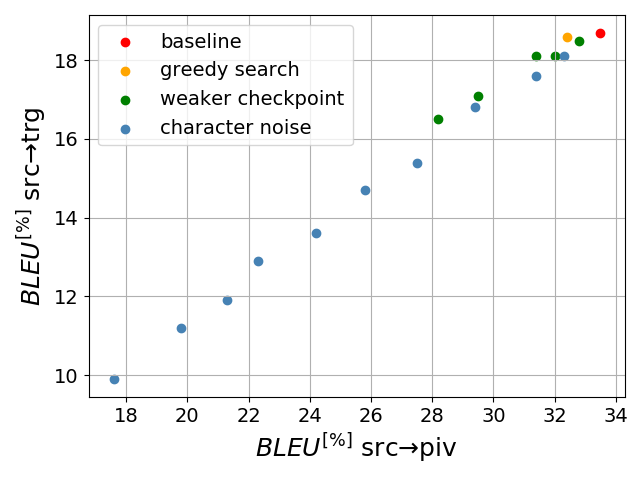}
    \caption{Impact of errors in the src$\to$piv model on the performance of the cascaded src$\to$trg system. 
    }
    \label{fig:error-propagation}
\end{figure}

Error propagation is a well-known problem of cascaded models.
In the following we investigate how significantly errors in one model influence the following models.
To this end, we monitor both the individual model performance and the end-to-end cascaded performance by running experiments on a three-way test set that consists of (source, pivot, target) triples.
For that purpose, we extract 3000 overlapping sentences from \texttt{NewsCommentary v14} for WMT French$\rightarrow$English and WMT English$\rightarrow$German to create a new test set that is disjoint with the training data. 
We train a 6-layer \lq base\rq{} Transformer for French$\rightarrow$English (src$\to$piv) and another for English$\rightarrow$German (piv$\to$trg).
In order to analyse the impact of disturbances and simulate errors in the French$\rightarrow$English system, we generate a weaker hypothesis by: 
\begin{itemize}
    \item Applying artificial character-level noise: With a probability of $p_{noise}$ each character in the decoded pivot hypothesis is replaced with a random character from the character set of the sentence 
    \item Using a weaker checkpoint than the baseline
    \item Reducing the beam size to 1 (greedy search)
\end{itemize}
By applying these procedures, we control the performance of the src$\to$piv model while maintaining a stable performance for the piv$\to$trg model. 
As is shown in Figure~\ref{fig:error-propagation}, the errors in the src$\to$piv model are actually deflated by the piv$\to$trg system, since a loss of 1.0 \BLEU{} in the src$\to$piv system results in only a drop of around 0.5 \BLEU{} for the cascaded src$\to$trg system.

Similarly, we conduct experiments in the other direction. 
By improving the quality of the prediction from the src$\to$piv model, we study the potential gain for the src$\to$trg task. 
For that purpose, we translate each source sentence to a 10-best list of pivot sentences. 
Using the pivot reference from the three-way test set we can select the single best hypothesis based on the sentence-level \BLEU{}

The sentence with the best \BLEU{} score among ten candidates is then passed to the piv$\to$trg model.
This cheating experiment results in an improvement of 6.2\% absolute \BLEU{} on the src$\to$piv model, which in turn however only results in 1.4\% absolute \BLEU{} improvement on the cascaded src$\to$trg model.
We conclude that (i) the piv$\to$trg models weakens both improvements and errors of the src$\to$piv model and (ii) the ambiguities in an src$\to$piv 10-best list hold room for an improvement of over 1.0 \BLEU{}.

\subsection{Effect of Training Data Size}
\label{sec:data-size}
To investigate how much the NAT-based integrated model quality depends on the training data size, we train our model on randomly sampled 50\%, 30\%, and 10\% selections of the original French$\rightarrow$German training corpus. To prevent overfitting on a small corpus, we increase the dropout rate to 0.3 compared to 0.1 on full French$\rightarrow$German corpus.
The Table~\ref{tab:fr-de-datasize} shows that when training on 10\% of the original data, the discrepancy between the best model performance is around 2.4\% \BLEU{}.
This setup simulates the data conditions of German$\rightarrow$Czech since the total amount of training sentences in German$\rightarrow$Czech corpus is around 10\% of the French$\rightarrow$German corpus. Based on our experimental results, we suppose that the integrated model needs some minimum amount of parallel src$\to$trg data to achieve the acceptable performance.

\begin{table}[ht]
\centering
\begin{tabular}{lc}
\hline
 data percentage & $\BLEU^{[\%]}$ \\ \hline
 100\%  & 21.5 \\
 50\%   & 21.0 \\
 30\%   & 20.6 \\
 10\%   & 19.1 \\
 \hline
\end{tabular}
\caption{French$\rightarrow$German dev set results using different training data partitions. The data percentage refers to the relative size of the training corpus comparing to the full French$\rightarrow$German training set. All experiments use the decoder states interface for NAT-based integrated training.}
\label{tab:fr-de-datasize}
\end{table}


\subsection{Effect of Model Pre-training}
In our experiments for the NAT-based integrated model, we solely rely on the models' pre-training, which means that instead of random initialization for the NAT-based integrated model components, we utilize the weights from the respective pre-trained models. In this section, we study the importance of model pre-training and its impact on the final model performance. For that purpose, we train the NAT-based integrated model with various initialization options.


\begin{figure}[ht]
    \centering
    \includegraphics[width=\linewidth]{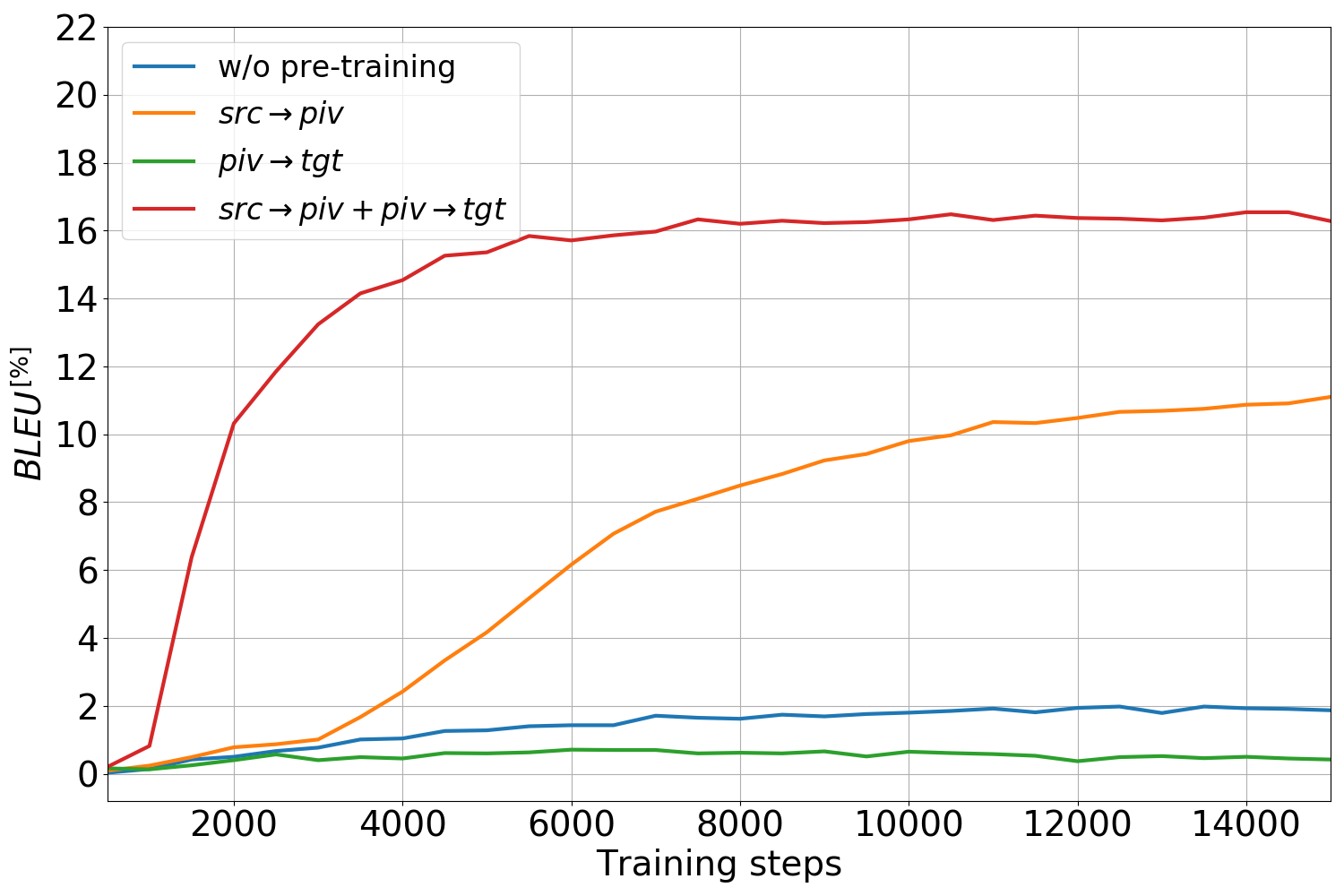}
    \caption{German$\rightarrow$Czech dev set results for different parameter pre-training schemes. src$\to$piv indicates that both $\Encoder_{s2p}$ and $\Decoder_{s2p}$ are pre-trained and all other parameters are randomly initialized. We use a similar notation for the other pre-training schemes. All experiments use the decoder states interface for NAT-based integrated training.}
    \label{fig:de-cs-intis}
\end{figure}

\begin{figure}[ht]
    \centering
    \includegraphics[width=\linewidth]{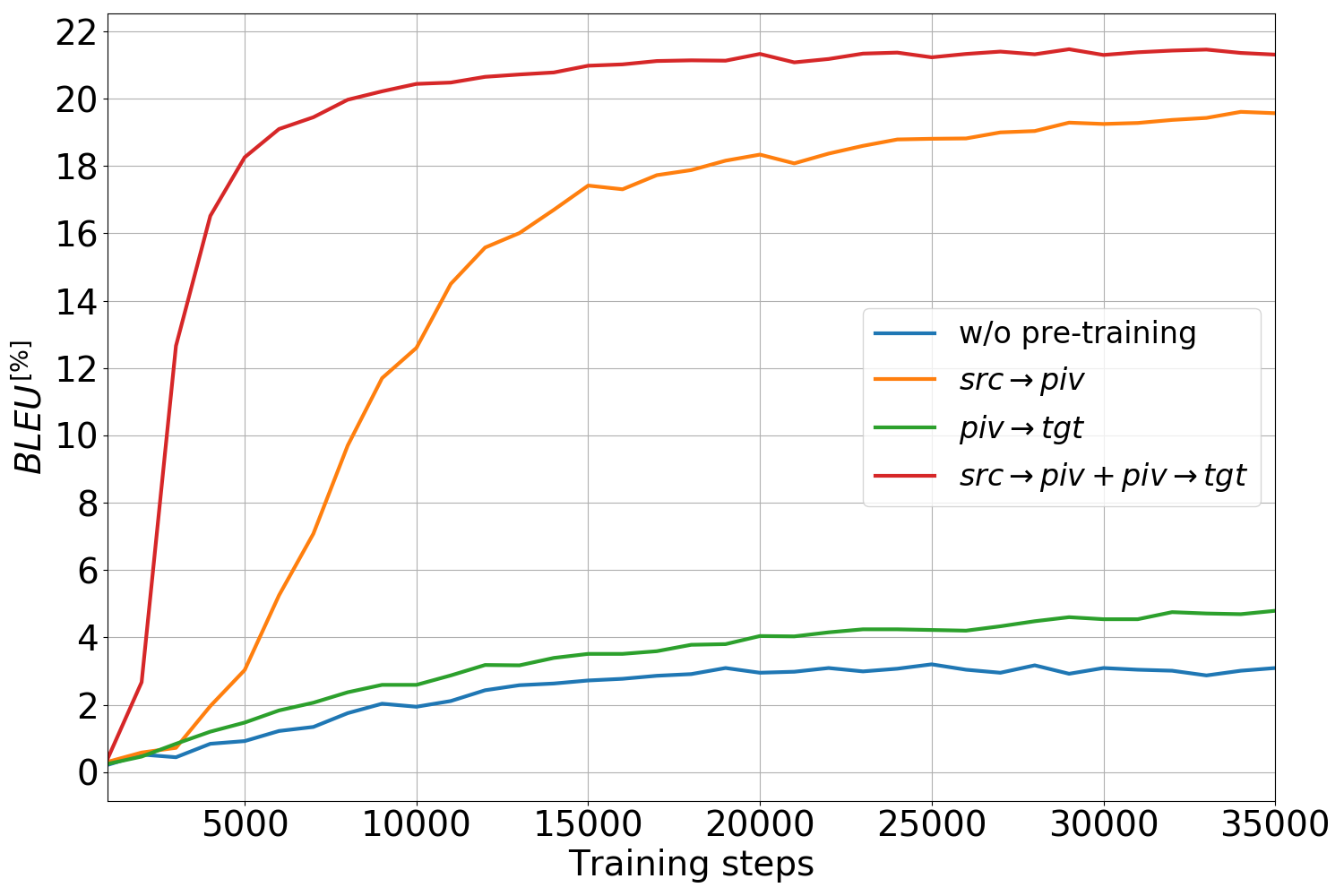}
    \caption{French$\rightarrow$German dev set results for different parameter pre-training schemes. All experiments use the decoder states interface for NAT-based integrated training.}
    \label{fig:fr-de-inits}
\end{figure}

Figure~\ref{fig:de-cs-intis} and Figure~\ref{fig:fr-de-inits} show that initialization of scr$\rightarrow$piv encoder and decoder is crucial for the final model performance.
Without initialization or with pre-training only piv$\to$trg encoder and decoder, it is impossible to train the end-to-end system. We see a similar trend while using the decoder posteriors interface.



\subsection{Length Modeling}
\label{sec:length-modeling}
Length modeling for the non-autoregressive decoder is one of the bottlenecks for our proposed NAT-based integrated model. The pivot sequence length has to be set in advance, and it can not be refined. In most of our experiments, we set the length of the intermediate sequence to be equal to the source sequence length both in training and test time. As a result, we do not fine-tune the length model using the src$\to$trg data. Moreover, the assumption that source length should match the pivot length does not hold for every language pair.
In Table~\ref{tab:length-results} we experiment with using different length estimates and report how it affects the end-to-end translation quality.

\begin{table}[ht]
\centering
\resizebox{\linewidth}{!}{%
\begin{tabular}{lccc}
\hline
 \multirow[b]{2}{*}{length source} & French$\rightarrow$German && German$\rightarrow$Czech \\ \hline
 & $\BLEU^{[\%]}$  && $\BLEU^{[\%]}$ \\ \hline
 random & 19.2 && 14.6\\
 source  & 21.6 && 16.8 \\
 target  & 18.9 && 16.5\\
 predicted & 21.3 && 17.2 \\
 \hline
\end{tabular}
}
\caption{Results for the different pivot length estimates on the dev set. Length source \texttt{random} refers to the length choice based on uniform distribution in the interval $[2,100)$. \texttt{predicted} refers to the usage of the CMLM length prediction component for length assignment. \texttt{source} and \texttt{target} indicate the length choice based on the source sequence or target sequence lengths. All experiments use the decoder posteriors interface for NAT-based integrated training.}
\label{tab:length-results}
\end{table}

The results show that better length modeling can lead to more than 2\% \BLEU{} improvements. However, for our experiments, we have not tried any sophisticated length prediction methods. We suppose that further exploration will be beneficial for the integrated model performance.

\subsection{Decoder Iterations}

The iterative refinement of the hypotheses by a non-autoregressive decoder plays an essential role in achieving better performance~\cite{ghazvininejad-etal-2019-mask, gu-2019-levenshtein}.
We observe that, the NA baseline with one decoder iteration of the src$\to$piv model results in 8.2 \BLEU{} on the French$\rightarrow$German development set, while five iterations of the same decoder yield 17.1 \BLEU{}. 
However, simply increasing the number of iterations during decoding with the integrated model does not lead to similar improvements.
Note that the output of the NA decoder is handed to an encoder, which a) more expressive than a softmax layer and b) is trained on the single-iteration output. This mismatch between training and decoding could be the reason why decoder iterations are not beneficial for the integrated model.
Additionally, we experimented with decoder iterations during training of the integrated model, but it breaks the gradient propagation. Although our initial experiments with the iterations have been unsuccessful, we think that they can be applied for training using such approaches as Gumbel-Softmax~\cite{jang-etal-2016-categorical}.

\subsection{Knowledge Distillation}
Sequence-level knowledge distillation (KD) ~\cite{kim-rush-2016-sequence} proved to be useful for the training of non-autoregressive models~\cite{zhou-etal-2020-understanding}. Although it improves the src$\to$piv model performance, our initial experiments show that KD results in a 0.1-0.3 \BLEU{} degradation on the integrated model.

\section{Conclusion}
In this work, we propose a novel architecture for the integrated training of cascaded models based on a non-autoregressive Transformer. We train the model on src$\to$piv, piv$\to$trg, and src$\to$trg data overcoming a drawback of conventional cascaded models.
Moreover, it provides a natural interface between two Transformer-based models and avoids unnecessary early decisions for intermediate representations. Our experimental results on the task of pivot-based machine translations show that the NAT-based integrated model outperforms the pivot baseline by up to 2.1\% \BLEU{} on WMT French$\rightarrow$German.

We analyze the integrated model and conclude that the src$\to$piv system is crucial for the final translation performance.
Further work is required to apply established NAT improvements to this new architecture, such as iterative decoding in the cascaded training and further experiments on knowledge distillation in the src$\to$piv pre-training, both of which show significant improvements in standalone systems~\cite{ghazvininejad-etal-2019-mask,GB18,gu-2019-levenshtein,zhou-etal-2020-understanding}.
Additionally, more sophisticated techniques for length modeling, such as an external length model or multiple length candidates, can be applied in the future to improve the quality of the pivot hypotheses.

Even though we test our cascaded architecture on the task for pivot-based machine translation, we can use the architecture in any application, where a combination of sequential models is beneficial.


\section*{Acknowledgments}
Authors affiliated with RWTH Aachen University have partially received funding from the European Research Council (ERC) (under the European Union's Horizon 2020 research and innovation programme, grant agreement No 694537, project ``SEQCLAS'') and eBay Inc. The work reflects only the authors' views, and none of the funding agencies is responsible for any use that may be made of the information it contains.

We would like to thank Yunsu Kim and Benedikt Hilmes for providing the autoregressive integrated model results.

\bibliographystyle{acl_natbib}
\bibliography{anthology,acl2021}


\end{document}